\documentclass[acmtog,timestamp]{acmart}

\usepackage{booktabs} 
\usepackage{subfig}
\usepackage[ruled]{algorithm2e} 

\setcopyright{none}

\usepackage{multicol}
\usepackage{amsmath}
\usepackage{svg}
\usepackage{graphicx}
\graphicspath{ {./images/} }
\usepackage{wrapfig}
\usepackage{caption}

\acmBooktitle{}

\newcommand{\Eref}[1]{Eq.~(\ref{#1})}
\newcommand{\Fref}[1]{Fig.~\ref{#1}}
\newcommand{\Sref}[1]{Sec.~\ref{#1}}

\begin{document}

\renewcommand\footnotetextcopyrightpermission[1]{}

\title{Generative Probabilistic Image Colorization}

\settopmatter{authorsperrow=4,printccs=false,printacmref=false}

\makeatletter

\author{Chie Furusawa}
\email{chie\_furusawa@dwango.co.jp}
\affiliation{%
  \institution{DWANGO Co., Ltd.}
  \country{Japan}}

\author{Shinya Kitaoka}
\email{shinya_kitaoka@dwango.co.jp}
\affiliation{%
  \institution{DWANGO Co., Ltd.}
  \country{Japan}}
  
\author{Michael Li}
\email{michael_li@dwango.co.jp}
\affiliation{%
  \institution{DWANGO Co., Ltd.}
  \country{Japan}}

\author{Yuri Odagiri}
\email{yuri_odagiri@dwango.co.jp}
\affiliation{%
  \institution{DWANGO Co., Ltd.}
  \country{Japan}}

\renewcommand{\shortauthors}{C. Furusawa et al.}

\begin{abstract}
We propose Generative Probabilistic Image Colorization, a diffusion-based generative process that trains a sequence of probabilistic models to reverse each step of noise corruption. Given a line-drawing image as input, our method suggests multiple candidate colorized images. Therefore, our method accounts for the ill-posed nature of the colorization problem. We conducted comprehensive experiments investigating the colorization of line-drawing images, report the influence of a score-based MCMC approach that corrects the marginal distribution of estimated samples, and further compare  different combinations of models and the similarity of their generated images. Despite using only a relatively small training dataset, we experimentally develop a method to generate multiple diverse colorization candidates which avoids mode collapse and does not require any additional constraints, losses, or re-training with alternative training conditions. Our proposed approach performed well not only on color-conditional image generation tasks using biased initial values, but also on some practical image completion and inpainting tasks.
\end{abstract}


\begin{teaserfigure}
  \centering
  \includegraphics[width=\textwidth]{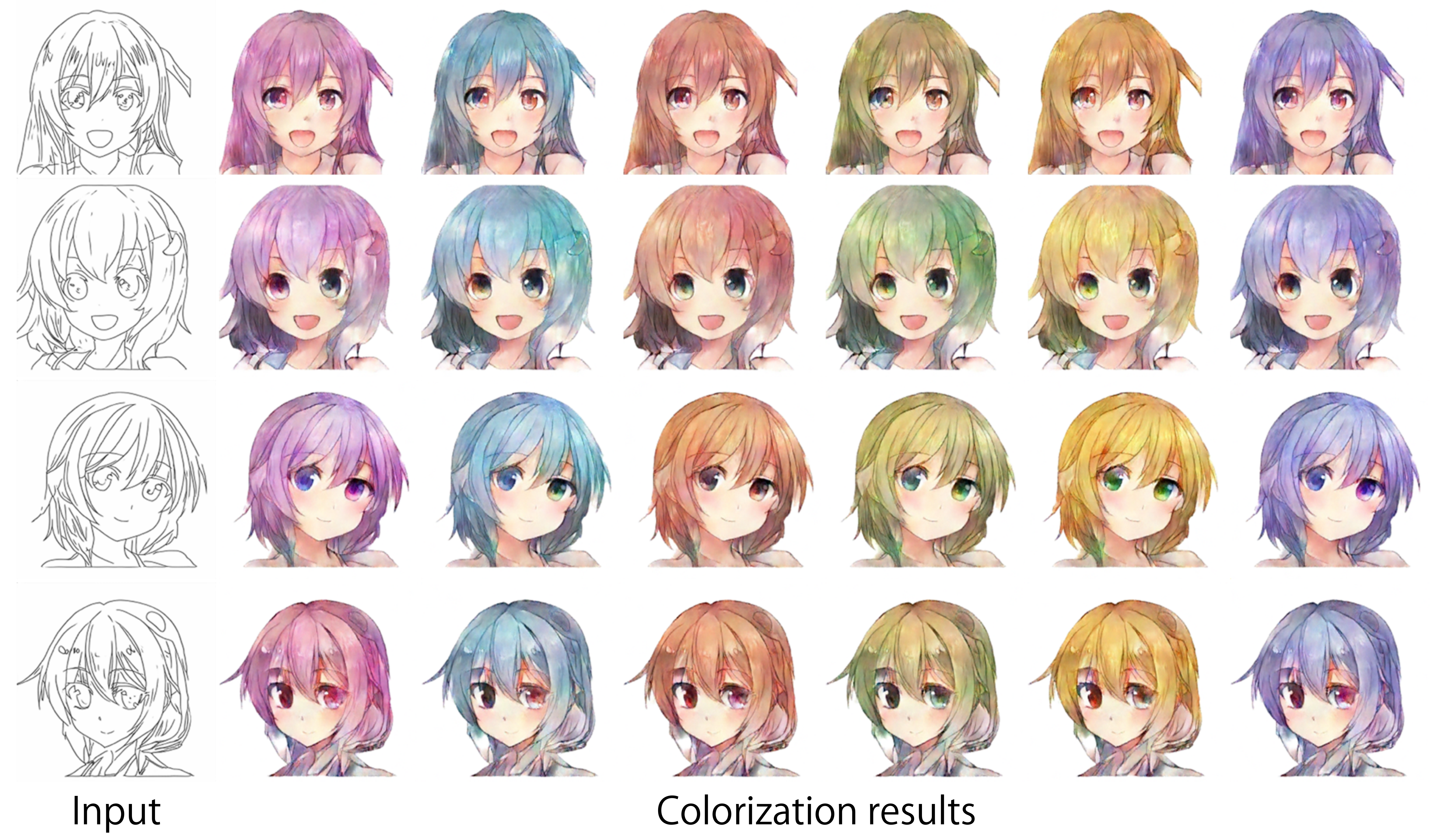}
   \caption{
   Colorization results of the proposed method. The leftmost column shows the input line-drawing images. Each row illustrates the colorization results from a single input image. Each colorization result represents an input image colorized to several different colors. Note that we provide neither manual interactive revisions, reference images, nor color dots for these images.
   }
   \label{fig:teaser}
\end{teaserfigure}

\maketitle

\section{Introduction}
There has long been widespread demand for image colorization methods for diverse applications in various fields. In fact, image colorization techniques have been actively investigated as a meaningful area of research for decades \cite{Tintfill, ComputerAssistedColoring, LazyBrush}. However, as yet no colorization methods have been developed that can satisfy the diverse aesthetic preferences of arbitrary people. It has thus far been considered impossible to determine a color composition for any arbitrary line drawing, as subjective preferences and aesthetic sensibilities are inherently complex, ambiguously defined, and subject to change over time.

Image colorization tasks may be divided into two types: tasks where an image must be colorized with specific colors (e.g. in anime or manga production processes) and tasks where the image does not have a known correct colorization (e.g. an original sketch created by a user). The latter task is an ill-posed problem, as there may be multiple possible solutions, and it is unknown if a solution is correct until the user sees it. One way to address this problem is to generate multiple colorization candidates for every input and allow the user to choose from them. Prior research on generating animations of multi-body simulations used sampling to suggest multiple different animations \cite{PlausibleSampling}.

Our main purpose is to suggest multiple and diverse colorization candidates from a single line-drawing.
Line drawings often do not have a predetermined color scheme. Manga is a good example. Manga characters are originally drawn in black-and-white. Cover artwork may show some of the characters in color, but not necessarily all of them, and the color scheme may be inconsistent. If the manga is adapted into an anime, color schemes must be chosen for all the characters. In this case, it would be useful to have a tool that can visualize multiple different colorization options, so the production team can choose the best one. In addition to manga, there are many other situations, such as industrial design, where the desired shape is known, but not the desired colors. These users could benefit from a system that can show them multiple different plausible colorizations, without requiring user hints. They would be able to look at the options to decide what colors they prefer. 

Our proposed method is meant for such situations, where users are choosing colors for the first time. We do not consider tasks in which the desired colorization is already known. In situations such as those we describe, the approach of choosing from multiple candidates is simple and useful. Other systems also suggest multiple predicted candidates to users. For example, predictive text systems and IMEs (Input Method Editors) show the user multiple candidates that the user then chooses from. Therefore, our method is useful during the initial stages of an illustration, to help choose a color scheme.

Deep learning-based colorization methods have recently exhibited high-quality results in a wide variety of applications. Building on these advancements, generative adversarial network (GAN) based image colorization approaches have also been developed. Some GAN-based methods have not been able to fully account for the ill-posed nature of the colorization problem. However, thanks to meticulously designed network architectures and careful parameter tuning, some multimodal image-to-image translation methods which use GANs can generate diverse colorizations. Alternatively, diffusion probabilistic models and flow-based models, which model the distributions for ill-posed problems, are other ways to generate multiple colorization candidates. In this paper, we especially experimented with whether the diffusion-based methods could generate diverse colorizations as well as GAN-based methods could.

We propose Generative Probabilistic Image Colorization, a diffusion-based generative process that trains a sequence of probabilistic models to reverse each step of noise corruption. Our proposed approach employs the connection between diffusion probabilistic models and denoising score matching with Langevin dynamics. To address the limitations of deep colorization methods, we employ diffusion probabilistic models, a class of latent variable models inspired by formulations used in nonequilibrium thermodynamics. This allows our model to suggest multiple and diverse colorization image candidates from a single input line-drawing image, without any additional constraints or losses. This is in contrast to conventional methods, which can only generate single or similar colorized images (see \Fref{fig:comparison}).

The main contributions of this study are as follows.
\begin{itemize}
\item In light of the ill-posed nature of the colorization problem, we suggest multiple colorized output candidates by employing diffusion probabilistic models.
\item Despite using only a relatively small training dataset, we experimentally develop a method to generate multiple and diverse colorization candidates stably which avoids mode collapse and does not require any additional constraints or losses.
\item We conducted comprehensive experiments on colorization of line-drawings, and the results show that our approach is able to perform well not only on color-conditional image generation using biased initial values, but also user-guided practical image synthesis tasks.
\end{itemize}

\section{Related Work}
\textbf{Rule-based methods: }
Colorization tools which do not rely on specific style features, such as homogeneous regions or pattern continuity, were proposed by \cite{LazyBrush}. In addition, colorization for use in manga based on color strokes was proposed based on classification schemes specific to manga textures \cite{Mangacolor}. These methods require input images with color strokes added by the user. 

\textbf{GAN-based methods: }
Some recent colorization methods have been proposed that train colorization models using GANs \cite{GAN,DCGAN}. GAN-based methods can generate diverse colorized images, the color compositions of which can vary greatly depending on the input. \cite{comicolorization, style2paintV2, userGuidedColorization2018} updated the GAN-based colorization approach to make use of a reference image, color dots, or both for colorization.  If the user used multiple different sets of reference color information (color dots or reference images), the user could get multiple and diverse colorized images from a single input. 
The automatic GAN-based colorization approach suffers from limitations in that the same input always produces the same colorization result, no matter how many times the colorization is performed \cite{pix2pix, bicyclegan}. To overcome these issues, other GAN-based methods employ meticulously designed network architectures with careful parameter tuning \cite{munit, MSGAN}. These methods tend to generate images which are diverse, but are missing parts, have faint colorization, or have color bleeding beyond the region. We experiment with a diffusion model for colorization and compare it with automatic GAN-based methods.

\textbf{Flow-based methods: }
Flow-based generative models are a conceptually attractive approach for generating multiple candidates. The exact latent-variable inference and exact log-likelihood are tractable, and both training and inference can be parallelized. 

\begin{algorithm}
\caption{Training}
    {1: \textbf{repeat} \\}
            {\ \ \ \ 2: $\mathbf{x}_{0} \sim q\left(\mathbf{x}_{0}\right)$ \\}
            {\ \ \ \ 3: $ \xi \sim \text {U}(0, 1)$ \\}
            {\ \ \ \ 4: $\bar{\alpha}_{t} \leftarrow \exp \left(-(\lambda \xi)^{2}\right)$ where $\lambda = 2.25$ \\}
            {\ \ \ \ 5: $\boldsymbol{\epsilon} \sim \mathcal{N}(\mathbf{0}, \mathbf{I})$ \\}
            {\ \ \ \ 6: Take gradient descent step on \\
                    {\ \ \ \ \ \ \ \ }
                    {$ \nabla_{\theta}|\boldsymbol{\epsilon}-\boldsymbol{\epsilon}_{\theta}\left(\sqrt{\bar{\alpha}_{t}} \mathbf{x}_{0}+\sqrt{1-\bar{\alpha}_{t}} \boldsymbol{\epsilon}, t,  \bar{\alpha}_{t}, \mathbf{x}_{0} |
                  \mathscr{L} (x)
                  \right)|
                    $ \\}
            }
    {7: \textbf{until} converged }
\label{alg:training}
\end{algorithm}

\vspace*{50pt}

\cite{NICE} and \cite{RealNVP} proposed a conditional normalizing flow architecture, and \cite{MAF} proposed a generative flow based on inverse autoregressive flows. \cite{Glow} proposed a simple type of generative flow, using an invertible 1 × 1 convolution. \cite{SRflow} applied a flow model for super-resolution tasks, which is also a class of ill-posed problems. While these methods can produce multiple predictions, they suffer from unstable training. For comparison with our proposed method, we adapted a network based on \cite{SRflow, Glow}, and then experimented with image colorization using the dataset used in our other experiments; however, the loss diverged during training. Moreover, the model produced by this method was relatively large.

\textbf{Diffusion-based methods: }
Diffusion probabilistic models can produce multiple candidates for a given task based on Langevin dynamics. \cite{VoiceGrad} employed dynamics with a score matching learning framework to achieve non-parallel any-to-many voice conversion, and \cite{DDPM} and \cite{SGM} showed that appropriate parameterization of diffusion models revealed an equivalence with denoising score matching over multiple noise levels during training and with annealed Langevin dynamics during sampling. The diffusion model demonstrated an ability to learn stably and to generate multiple predictions with different sampling. We therefore applied diffusion probabilistic models to the colorization task.

In summary, rule-based methods do not present multiple candidates as solutions to the ill-posed problem of image colorization. Furthermore, flow-based methods cannot be trained stably. We examine diffusion probabilistic models for obtaining multiple candidates for image colorization and compare our results to GAN-based methods experimentally.

\textbf{Other methods: }
Other methods which use neural networks have been proposed that learn from copious amounts of color illustrations \cite{Scribbler, userGuidedColorization2017, exmaplerBasedColorization2018, userGuidedColorization2021}. These methods are user-guided. Our purpose is to get multiple diverse colorization candidates at once without user guidance, therefore, in this paper, we exclude these methods from comparison. 

\begin{algorithm}
{1: $\mathbf{x}_{T} \sim \mathcal{N}(\mathbf{0}, \mathbf{I})$ \\}
{2: \textbf{for} t = T, . . . , 1 \textbf{do} \\}
    \textit{\ \ \ \ \# Predictor \\}
    {\ \ \ \ 3: $\mathbf{z} \sim \mathcal{N}(\mathbf{0}, \mathbf{I}) \text { if } t>1, \text { else } \mathbf{z}=\mathbf{0} $ \\}
    {\ \ \ \ 4: $ \mathbf{x}_{t-1}=\frac{1}{\sqrt{\alpha_{t}}}\left(\mathbf{x}_{t}-\frac{1-\alpha_{t}}{\sqrt{1-\bar{\alpha}_{t}}} \boldsymbol{\epsilon}_{\theta}\left(\mathbf{x}_{t}, t\right)\right)+\sigma_{t} \mathbf{z} $ \\}
    \textit{\ \ \ \ \# Corrector \\}
    {\ \ \ \ 5: \textbf{for} n = 1, . . . , M \textbf{do} \\}
        {\ \ \ \ \ \ \ \ }
        {6: $\mathbf{z} \sim \mathcal{N}(\mathbf{0}, \mathbf{I})$ \\}
        {\ \ \ \ \ \ \ \ }
        {7: $\epsilon_{t} \leftarrow 2 \alpha_{t} \frac{|\mathrm{z}|^{2}}{\left|\epsilon_{\theta}\left(\mathrm{x}_{t}, \bar{\alpha}_{t}\right)\right|^{2}}$ \\}
        {\ \ \ \ \ \ \ \ }
        {8: $\mathbf{x}_{t} \leftarrow \mathbf{x}_{t} -\epsilon_{t} \frac{\epsilon_{\theta}\left(\mathbf{x}_{t}, t\right)}{\sqrt{1-\bar{\alpha}_{t}}} \left(\mathbf{x}_{t}, t\right)+\sqrt{2 \epsilon_{t}} \mathbf{z}$ \\}
{9: \textbf{end for} \\}
{10: \textbf{return} $\mathbf{x}_{0}$}
\caption{Sampling}
\label{alg:sampling}
\end{algorithm}

\newpage

\begin{figure*}[!htb]
\minipage[t]{0.32\textwidth}
  \includegraphics[width=\linewidth]{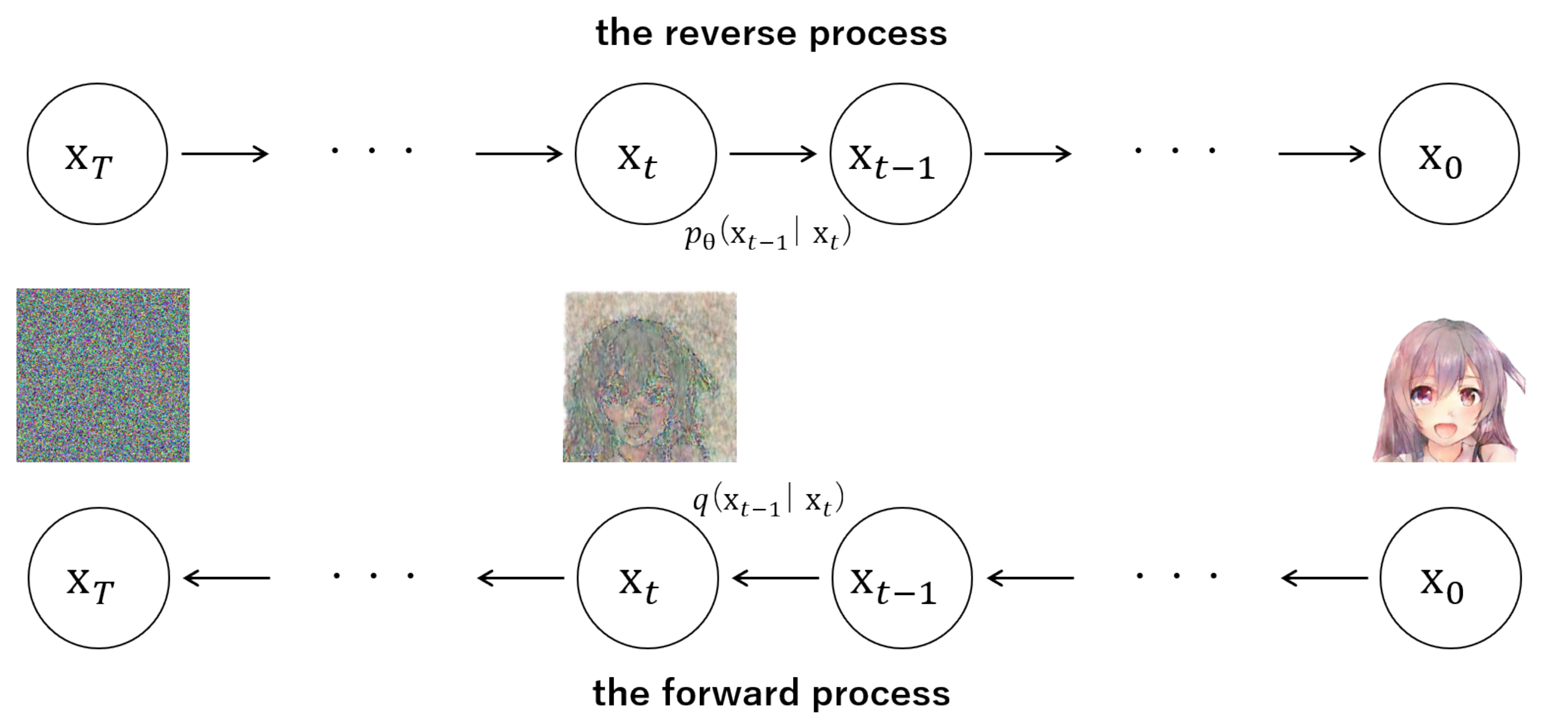}
  \caption{Graphical model of Generative Probabilistic Image Colorization. In the forward process, a Markov chain gradually adds noise to the data in the opposite direction of sampling until the signal is destroyed, while in the reverse process, the chain transforms the noise into the objective image with learned Gaussian transitions.}
  \label{fig:process}
\endminipage\hfill
\minipage[t]{0.32\textwidth}
  \includegraphics[width=\linewidth]{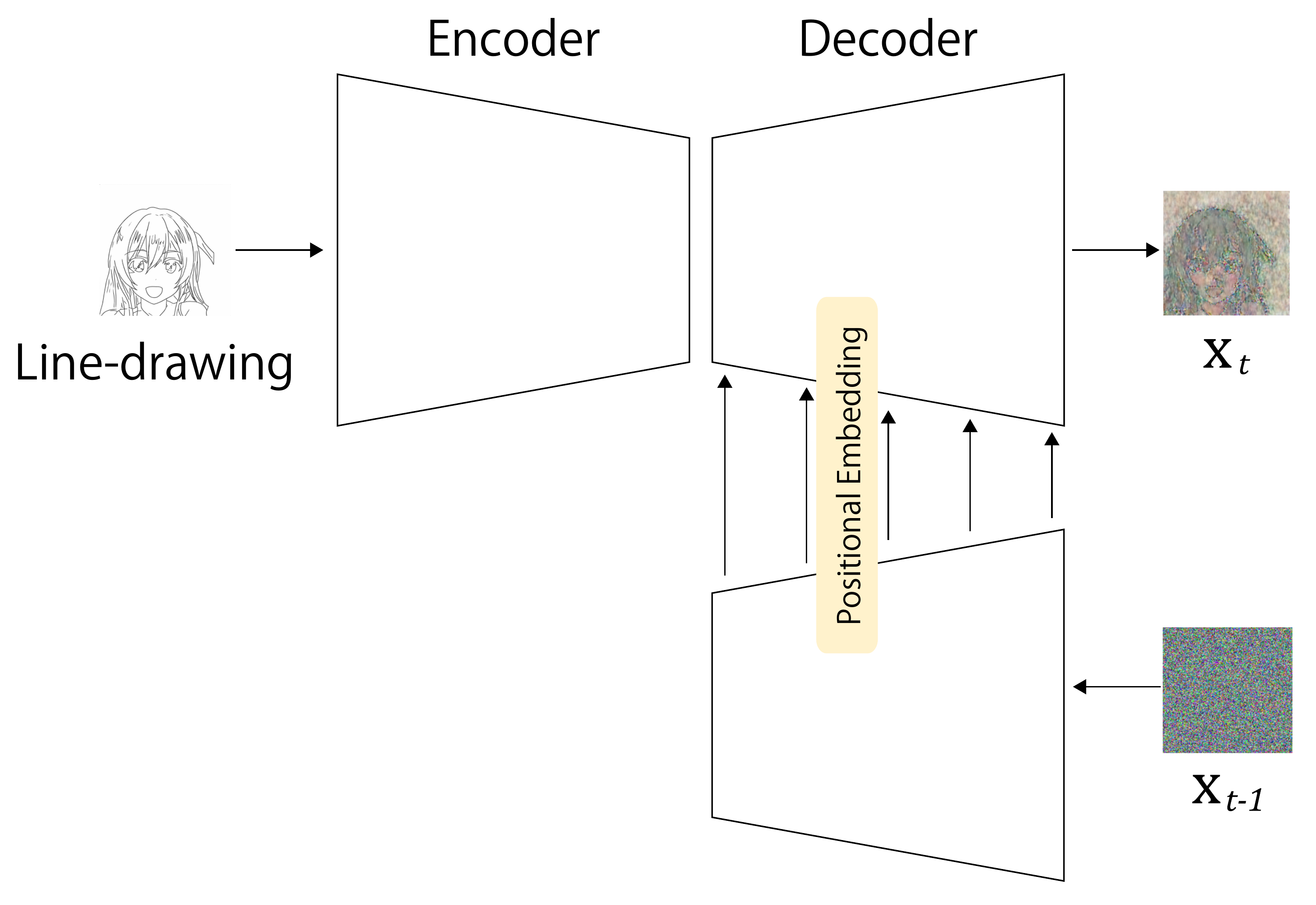}
  \caption{Network architecture of the proposed method. Our approach utilizes an encoder-decoder network, and we additionally design a positional embedding architecture to encode $\bar{\alpha}_{t}$ at each timestep. A line-drawing image is input to the network and is then encoded, and then the features from the encoder are decoded along with the positional embeddings from the previous step's output.
  }
  \label{fig:network_simple}
\endminipage\hfill
\minipage[t]{0.32\textwidth}%
  \includegraphics[width=\linewidth]{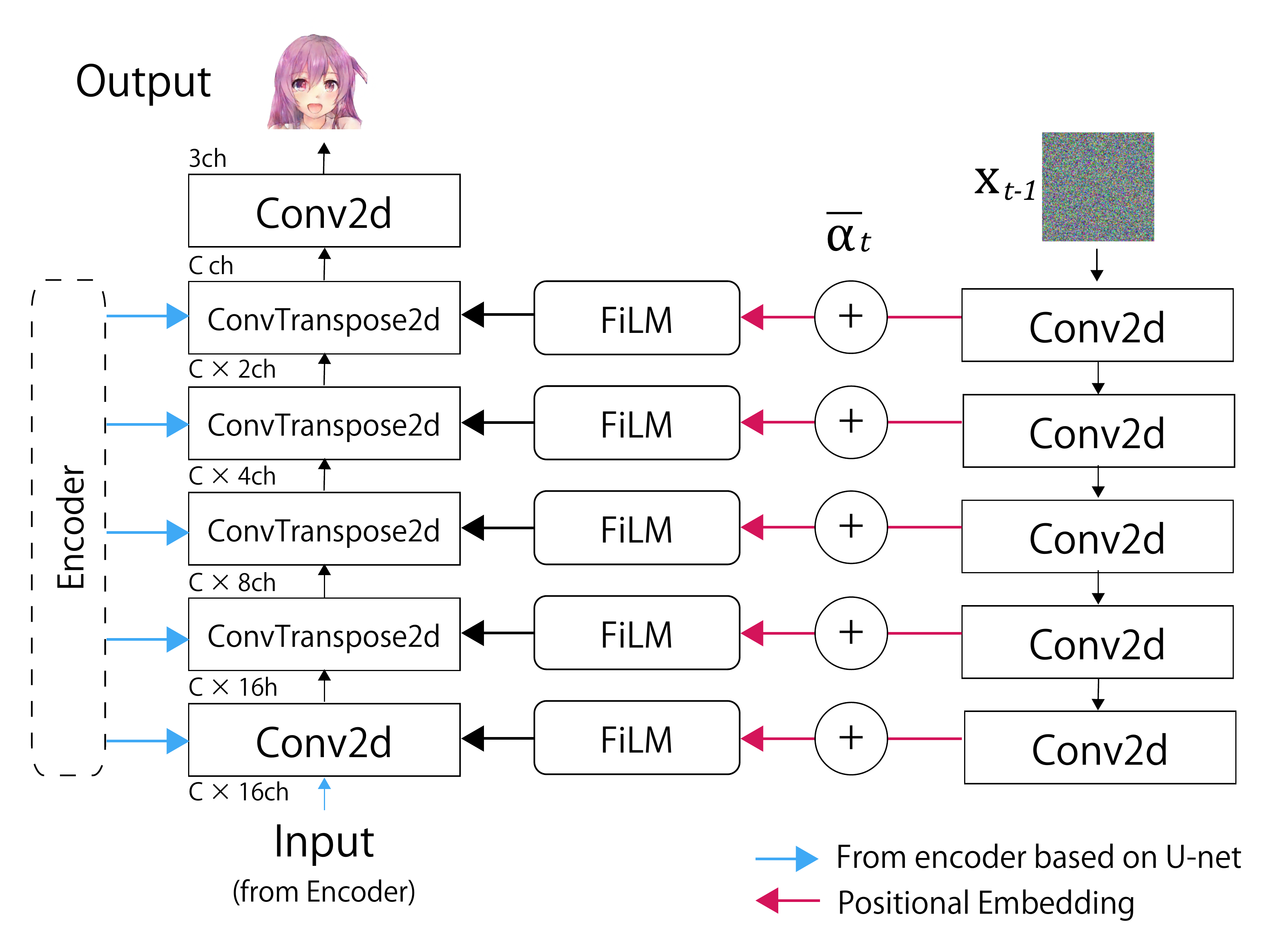}
  \caption{Architecture of our decoder. $C$ is a parameter that dictates the number of channels. We use a U-Net backbone for the encoder. The features are decoded with the features from $\mathbf{x}_{t-1}$ and positional embeddings obtained by passing $\bar{\alpha}_{t}$ through a 5-layer perceptron with Mish \cite{Mish}. The FiLM module \cite{WaneGrad} produces both scale and bias vectors from a given input that it then applies to the outputs of the ConvTranspose layers.
}
  \label{fig:network_main}
\endminipage
\end{figure*}

\section{Method}
\subsection{Diffusion Probabilistic Models}
A diffusion probabilistic model is composed of a variational Markov chain. In the forward process, the Markov chain gradually adds noise to the data in the opposite direction of sampling until the signal is destroyed, while in the reverse process, the chain transforms the noise into the objective image with learned Gaussian transitions, as depicted in \Fref{fig:process}. 

Diffusion models assume the data distribution can be modeled as

\begin{equation}
    p_{\theta}\left(\mathbf{x}_{0}\right):=\int p_{\theta}\left(\mathbf{x}_{0: T}\right) d \mathbf{x}_{1: T},
\end{equation}
where $ x_1 $ , . . . , $ x_T $ are latent variables, each of which is of the same dimension as the data $ x_0 $, and T is the number of iterations of the generation processes.

The reverse process $ p_{\theta}\left(\mathbf{x}_{0: T}\right) $ is defined as a Markov chain with learned Gaussian transitions starting at $ p\left(\mathbf{x}_{T}\right)=\mathcal{N}\left(\mathbf{x}_{T} ; \mathbf{0}, \mathbf{I}\right)
$ as 
 
\begin{equation}
p_{\theta}\left(\mathbf{x}_{0: T}\right):=p\left(\mathbf{x}_{T}\right) \prod_{t-1}^{T} p_{\theta}\left(\mathbf{x}_{t-1} \mid \mathbf{x}_{t}\right),
\end{equation}
where each iteration proceeds to the next time step as follows.

\begin{equation}
p_{\theta}\left(\mathbf{x}_{t-1} \mid \mathbf{x}_{t}\right):=\mathcal{N}\left(\mathbf{x}_{t-1} ; \boldsymbol{\mu}_{\theta}\left(\mathbf{x}_{t}, t\right), \mathbf{\Sigma}_{\theta}\left(\mathbf{x}_{t}, t\right)\right).
\end{equation}
The forward process $ q\left(\mathbf{x}_{1: T} \mid \mathbf{x}_{0}\right) $ is defined as

\begin{equation}
q\left(\mathbf{x}_{1: T} \mid \mathbf{x}_{0}\right):=\prod_{t=1}^{T} q\left(\mathbf{x}_{t} \mid \mathbf{x}_{t-1}\right),
\end{equation}
where each iteration adds noise arbitrarily following a Gaussian distribution formulated as 

\begin{equation}
q\left(\mathbf{x}_{t} \mid \mathbf{x}_{t-1}\right):=\mathcal{N}\left(\mathbf{x}_{t} ; \sqrt{1-\beta_{t}} \mathbf{x}_{t-1}, \beta_{t} \mathbf{I}\right),
\label{eq:forward}
\end{equation}
under a variance noise schedule $\beta_{1}, \ldots, \beta_{T}$.  Using the notation $ \alpha_{t}:=1-\beta_{t} $ and $ \bar{\alpha}_{t}:=\prod_{s=1}^{t} \alpha_{s} $, we can write \Eref{eq:forward} in closed form for a sample $\mathbf{x}_{t}$ at an arbitrary timestep $t$ as

\begin{equation}
q\left(\mathbf{x}_{t} \mid \mathbf{x}_{0}\right)=\mathcal{N}\left(\mathbf{x}_{t} ; \sqrt{\bar{\alpha}_{t}} \mathbf{x}_{0},\left(1-\bar{\alpha}_{t}\right) \mathbf{I}\right).
\end{equation}

During training, we maximize the following evidence lower bound instead of the log likelihood $ \mathbb{E}\left[\log p_{\theta}\left(\mathbf{x}_{0}\right)\right] $.

\begin{equation}
\begin{aligned}
\mathbb{E}\left[\log p_{\theta}\left(\mathbf{x}_{0}\right)\right] & \geq \mathbb{E}_{q}\left[\log \frac{p_{\theta}\left(\mathbf{x}_{0: T}\right)}{q\left(\mathbf{x}_{1: T} \mid \mathbf{x}_{0}\right)}\right] \\
&=\mathbb{E}_{q}\left[\log p\left(\mathbf{x}_{T}\right)+\sum_{t \geq 1} \log \frac{p_{\theta}\left(\mathbf{x}_{t-1} \mid \mathbf{x}_{t}\right)}{q\left(\mathbf{x}_{t} \mid \mathbf{x}_{t-1}\right)}\right]=: L. 
\label{eq:elbo}
\end{aligned}
\end{equation}

\cite{DDPM} and \cite{WaneGrad} found that the $L$ in \Eref{eq:elbo} could be simplified in a way that is beneficial to sample quality and easier to implement, as follows:

\begin{equation}
L_{\text {simple }}(\theta):=\mathbb{E}_{t, \mathbf{x}_{0}, \epsilon}\left[\left|\boldsymbol{\epsilon}-\boldsymbol{\epsilon}_{\theta}\left(\sqrt{\bar{\alpha}_{t}} \mathbf{x}_{0}+\sqrt{1-\bar{\alpha}_{t}} \boldsymbol{\epsilon}, t\right)\right|\right],
\end{equation}
where $\boldsymbol{\epsilon} \sim \mathcal{N}(\mathbf{0}, \mathbf{I})$, and $\boldsymbol{\epsilon}_{\theta}$ is a function approximator that the network learns. Furthermore, the sampling procedure is performed according to \cite{DDPM} as

\begin{equation}
\mathbf{x}_{t-1}=\frac{1}{\sqrt{\alpha_{t}}}\left(\mathbf{x}_{t}-\frac{\beta_{t}}{\sqrt{1-\bar{\alpha}_{t}}} \boldsymbol{\epsilon}_{\theta}\left(\mathbf{x}_{t}, t\right)\right)+\sigma_{t} \mathbf{z},
\label{eq:predictor}
\end{equation}
where $\sigma_{t}^{2}=\frac{1-\bar{\alpha}_{t-1}}{1-\bar{\alpha}_{t}} \beta_{t}$, and $\mathbf{z} \sim \mathcal{N}(\mathbf{0}, \mathbf{I})$.

\subsection{Network Architecture}
\Fref{fig:network_simple} shows the architecture of our proposed network. Our approach utilizes an encoder-decoder network, and we additionally design a positional embedding architecture to encode $\bar{\alpha}_{t}$ at each timestep. \Fref{fig:network_simple} shows that a line-drawing image is input to the network and is then encoded, and then the features from the encoder are decoded along with the positional embeddings from the previous step's output.

The state of the previous timestep $\mathbf{x}_{t-1}$ and $\bar{\alpha}_{t}$ are concatenated. $\mathbf{x}_{t-1}$ is fed through the block of Conv2D layers, and we use the Fourier features of $\bar{\alpha}_{t}$, which are extracted following \cite{FourierFeatures}. This process maps the input into a higher-dimensional feature space before passing it through the network, as follows. 

\begin{equation}
\gamma(\bar{\alpha}_{t}) = [\cos(2\pi\mathbf{b}^{\top} \bar{\alpha}_{t}), \sin(2\pi\mathbf{b}^{\top} \bar{\alpha}_{t})]^{\top},
\label{eq:fourier_features}
\end{equation}
where $\mathbf{b} \in \mathbb{R}^{D}$ is a random Gaussian vector. These Fourier features are embedded in the decoder as depicted in \Fref{fig:network_main}. 

As shown in \Fref{fig:network_main}, our network utilizes an encoder based on U-Net \cite{U_net}. The encoded features of a line-drawing image are decoded with the features from $\mathbf{x}_{t-1}$ and positional embeddings obtained by passing $\bar{\alpha}_{t}$ through a 5-layer perceptron with a self-regularized non-monotonic activation function, Mish \cite{Mish}. The FiLM (feature-wise linear modulation) module \cite{WaneGrad} produces both scale and bias vectors from a given input that it then applies to the outputs of the ConvTranspose layers. The model produces an output ${x}_{t}$ at each iteration, which can be interpreted as an update of one timestep to ${x}_{t-1}$. 
 
\subsection{Sampling of the variance schedule}
\begin{wrapfigure}{L}{0.5\linewidth}
\centering
  \includegraphics[width=\linewidth]{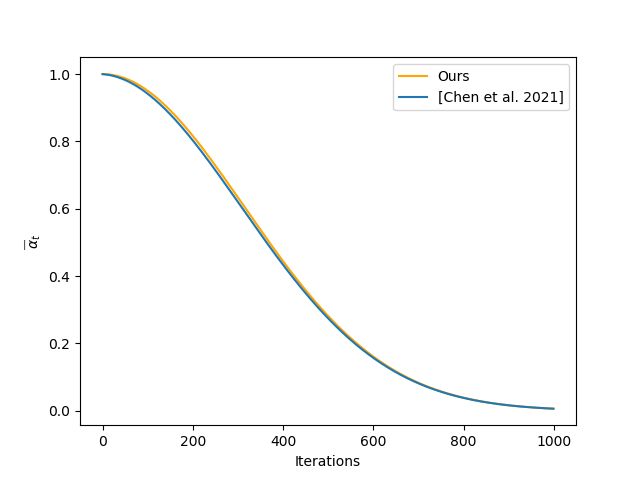}
  \caption{
  We use a simple hierarchical sampling that mimics discrete sampling. The orange line closely approximates the blue line.
  }
  \label{fig:sampled_alpha_bar}
\end{wrapfigure}

For simplicity, we used an approximated $ \bar{\alpha}_{t}$ in training. Following the approach of \cite{DDPM}, we gradually added Gaussian noise to the data according to the value $ \bar{\alpha}_{t} $, which was derived from the variance schedule, and we applied the direct sampling procedure presented in \cite{WaneGrad}. Additionally, we used a simple hierarchical sampling that mimics discrete sampling, as follows:

\begin{equation}
\bar{\alpha}_{t} \simeq \exp \left(-(\lambda\xi)^{2}\right),
\end{equation}
where ${\xi}$ is a random number from a uniform distribution on the interval [0, 1), and we set $ \lambda $ to 2.25. We determine the value of the hyperparameter $\lambda$ based on the total number of timesteps. \Fref{fig:sampled_alpha_bar} shows the resulting distributions obtained by the method used in \cite{WaneGrad} and by our proposed method.

In \Fref{fig:sampled_alpha_bar}, the orange line is the sampled variance schedule used in training, and the blue line is the variance schedule used by \cite{WaneGrad}. The orange line closely approximates the blue line; therefore, we considered our sampling of $ \bar{\alpha}_{t} $ to be sufficient and used this sampled $ \bar{\alpha}_{t} $ to train relatively simply.

\subsection{Predictor-corrector framework}
\label{sec:predictor_corrector}
Our method employs a predictor-corrector framework when generating in the reverse process as proposed by \cite{SGM} in order to generate a prediction. They explained that a predictor-corrector framework is useful to correct errors. \Eref{eq:predictor} corresponds to the “predictor”, and they added a "corrector" in the reverse process, a score-based Markov chain Monte Carlo (MCMC) approach that corrects the marginal distribution of the estimated sample \cite{CorrelationFuncs, MCMC}.
We derive the relationship between the score function $ s_{\theta} $ \cite{CorrelationFuncs, MCMC} and $ \epsilon_{\theta} $, and then add the corrector in the reverse process using the following equation. 

\begin{equation}
\begin{aligned}
s_{\theta}\left(\mathrm{x}_{t}, t\right)=\nabla_{\mathrm{x}_{t}} \ln \frac{1}{\sqrt{2 \pi\left(1-\bar{\alpha}_{t}\right)}} \exp \left(-\frac{\left(x_{t}-\sqrt{\bar{\alpha}_{t}} x_{0}\right)^{2}}{2\left(1-\bar{\alpha}_{t}\right)}\right)=-\frac{\epsilon_{\theta}\left(\mathrm{x}_{t}, t\right)}{\sqrt{1-\bar{\alpha}_{t}}}.
\end{aligned}
\label{eq:s_upsilon_relation}
\end{equation}

where $ s_{\theta} $ is a time-dependent score-based model.

\subsection{Bias of initial values}
To obtain various candidates, we perform color-conditional image generation, using the biased color condition $\textbf{x}_{T}$ as the initial value in the generation process, as follows:

\begin{equation}
\begin{array}{c}
\textbf{x}_{T}= \sqrt{\bar{\alpha}_{T}} \mathbf{V} + \sqrt{1-\bar{\alpha}_{T}} \epsilon ,
\label{eq:inital_bias}
\end{array}
\end{equation}
where $\bar{\alpha}_{T} =\exp\left(-\lambda^{2}\right)$, $\lambda = 2.25$, $\epsilon \sim \mathcal{N}(0, I)$, and $\textbf{V} \in \mathbb{R}^{W\times H\times 3}$ is a tensor filled with the biased RGB value, which is equivalent to a single-color image. We experimentally found that this color bias could roughly determine the output colors. We used this bias in every experiment in this study, and set $\textbf{V}$ to different color values in the reverse process. The biased initial value provided different colorizations (see \Fref{fig:teaser} and \Fref{fig:comparison}).

Finally, we describe the algorithms used in training and sampling in Algorithm 1 and Algorithm 2, respectively.

\subsection{Datasets}

\begin{figure}[t]
  \centering
  \includegraphics[width=\linewidth]{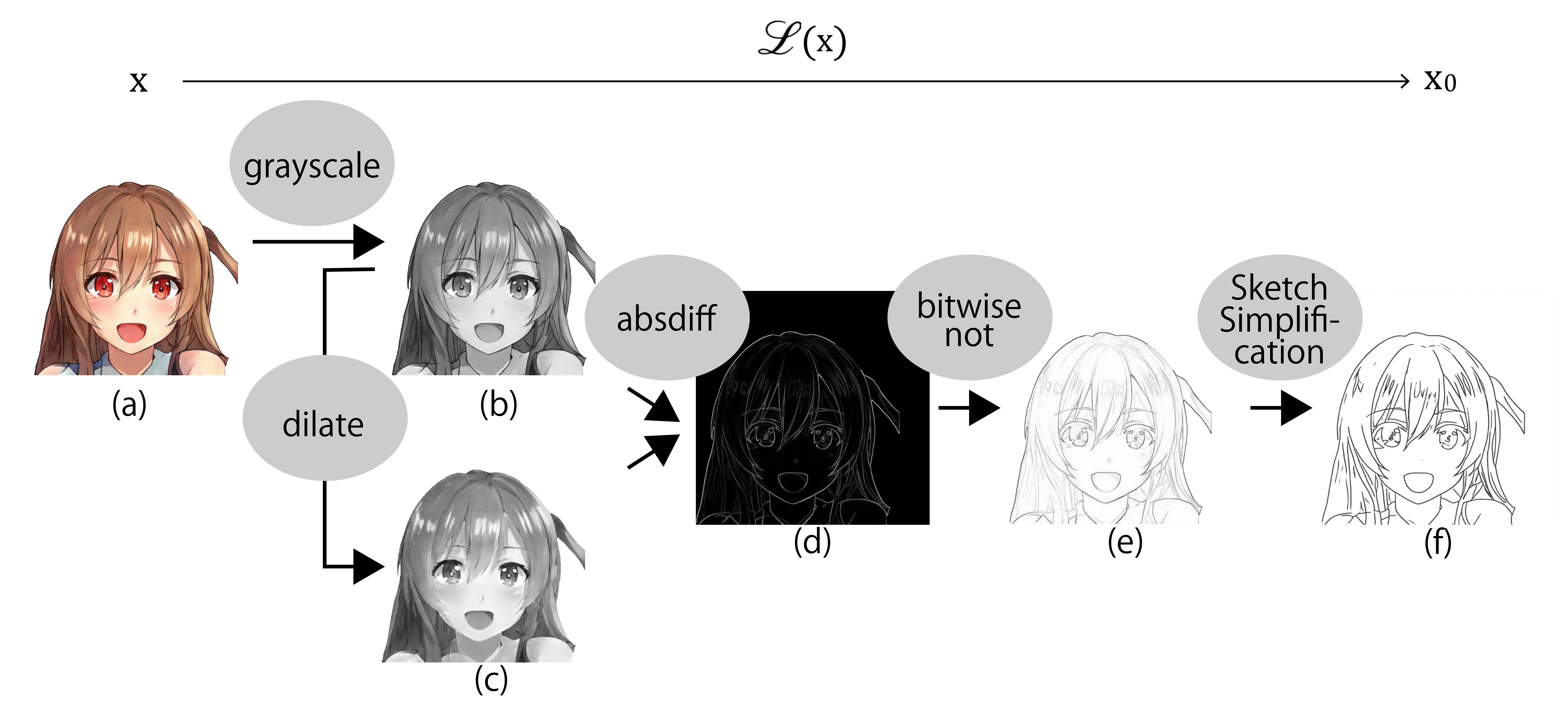}
  \caption{
(a) Input image. (b) Load (a) in grayscale. (c) Perform dilation on (b). (d) Difference of images (b) and (c). (e) Bitwise-inverted image. (f) Output of simplification performed using \cite{SketchSimplification}. We define this sequential processes for the generation of line-drawings as a function $ \mathscr{L}({x})$, and we use (a) and (f) as training pairs.
}
  \label{fig:datasets}
\end{figure}

To train our proposed model, we constructed character datasets based on \textit{Artbreeder} \footnote{https://www.artbreeder.com/}, which posts large numbers of illustrations produced by GAN. We crawled the Artbreeder website and collected 450 color images for training, and then we performed preprocessing as follows. 
(1) We removed the backgrounds from the character images, because the input images had meaningless artificially generated backgrounds. 
(2) We transformed those images into line-drawing images that served as input data for our method. We loaded the original images in grayscale. Next, we performed dilation on the grayscale input. Then we computed the difference between the dilated image and the grayscale image and bitwise inverted the result to produce a sketch-like image. Finally, this sketch-like image was input into the image simplification model proposed by \cite{SketchSimplification}. \Fref{fig:datasets} illustrates this process.
(3) We resized each image to 256 × 256. 
We define this sequential process for the generation of line-drawings as a function $\mathscr{L}({x})$, that is, ${x}_{0} = \mathscr{L}({x})$.

\section{Experiments}
\begin{figure*}[t]
\minipage[t]{0.47\textwidth}
  \includegraphics[width=\linewidth]{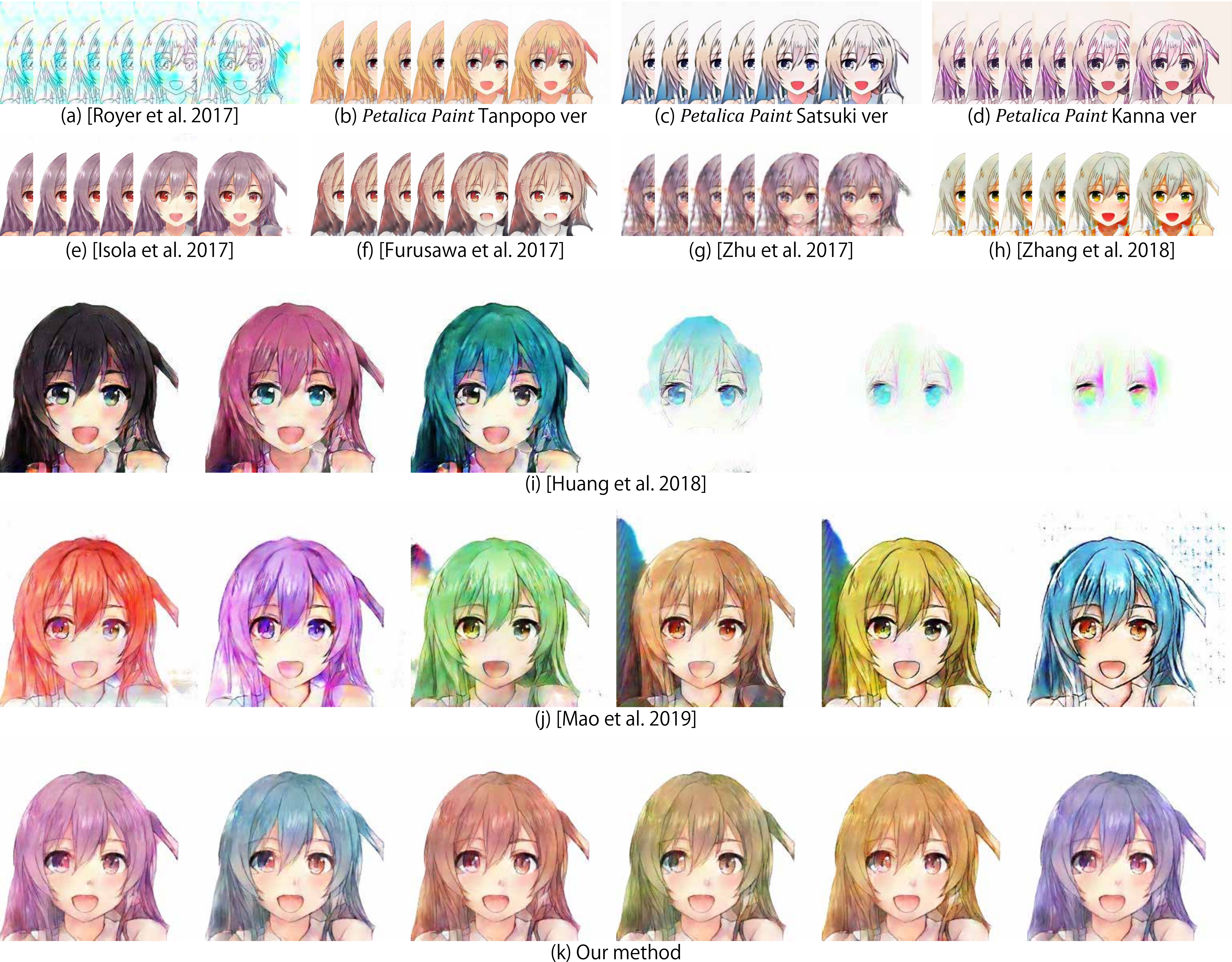}
  \caption{
  Comparison with state-of-the-art image colorization methods. 
  (a) \cite{PIC}. 
  (b) \textit{Petalica Paint} Tanpopo ver.
  (c) \textit{Petalica Paint} Satsuki ver.
  (d) \textit{Petalica Paint} Kanna ver.
  (e) \cite{pix2pix}. 
  (f) \cite{comicolorization}.
  (g) \cite{bicyclegan}
  (h) \cite{style2paintV2}.
  (i) \cite{munit}
  (j) \cite{MSGAN}.
  (k) Our method. 
  Our method plausibly colorized each region of the input image; additionally, our approach reliably generates multiple candidates that vary in appearance without any reference images or color dots. In contrast, other methods generate some poor results or only similar colorization results.
}
  \label{fig:comparison}
\endminipage\hfill
\minipage[t]{0.47\textwidth}
  \includegraphics[width=\linewidth]{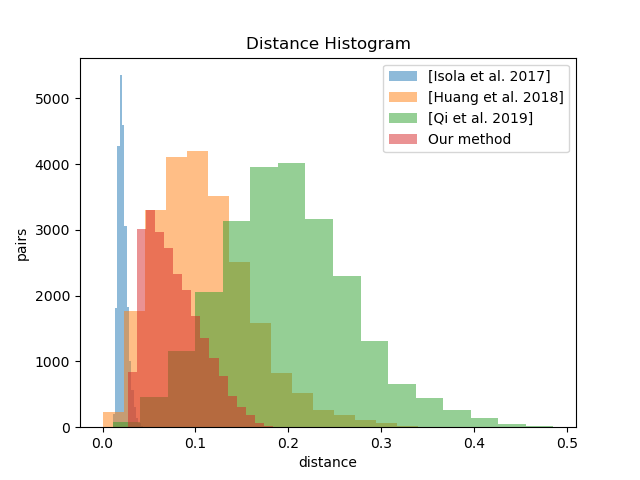}
  \caption{
Comparison of image similarity, as computed based on \cite{PerceptualSimilarity}. Blue bins are \cite{pix2pix}, orange bins are \cite{munit}, green bins are \cite{MSGAN}, and red bins are our method.
The distribution of the blue bins, corresponding to \cite{pix2pix}, is narrow and near the origin. The distributions of the orange and green bins, corresponding to \cite{munit} and \cite{MSGAN}, look wide and far from the origin, but these methods have some quality issues, as explained in \Sref{sec:results}. Therefore, the greater distances of the green and orange bins are due to poor results. On the other hand, our method produces reasonably varied images without sacrificing image quality and stability.
}
  \label{fig:comparison_dists}
\endminipage\hfill
\end{figure*}

\subsection{Experimental Conditions}
We used one million epochs to train our proposed model because the model trained with one million epochs had the lowest validation loss. We used Ranger, a synergistic optimizer using Rectified Adam (RAdam) and LookAhead in a combined codebase \cite{Radam, LookAhead, GradientCentra}. The parameters for this optimizer were as follows: the learning rate was $10^{-4} $, alpha was $0.5$, k was $6$, N\_sma\_threshhold was $5$, betas were $(0.95, 0.999)$, eps was $10^{-5}$, and weight\_decay was $0$.

\subsection{Results}
\label{sec:results}
\Fref{fig:teaser} shows inputs and the colorization results produced by our method. Our model generates multiple and diverse colorization candidates from a single input. We used two types of models with different numbers of channels in their convolutional layers. The number of channels is dictated by the parameter $C$ as in \Fref{fig:network_main}. Our two types of models use $ C=8 $ and $ C=32 $. We set the total number of iterations T of the reverse process to 1000. Over the 1000 iterations, we first used a model with $C=8$ for 960 iterations, then switched to a model with $C=32$ for the remaining 40. We discuss the number of iterations and the relations between the number of channels and the colorization results in \Sref{sec:channels}.

\Fref{fig:comparison} illustrates a comparison of our method with other colorization approaches. The input of \Fref{fig:comparison} is the same as the top row of \Fref{fig:teaser}. Our method plausibly colorized each region of the input image; additionally, our method (\Fref{fig:comparison}(k)) and the methods in \Fref{fig:comparison}(i) and (j) reliably generated multiple candidates that vary in appearance.In contrast, \Fref{fig:comparison}(a-h) generated unfavorable results or only similar colorization results. 
\Fref{fig:comparison}(a) shows the results generated by \cite{PIC}, which was trained using their source code and our datasets; their method was unable to colorize each region. \Fref{fig:comparison}(b)-(h) show that \textit{Petalica Paint} \footnote{https://petalica-paint.pixiv.dev}, \cite{pix2pix}, \cite{comicolorization}, \cite{bicyclegan}, and \cite{style2paintV2} could only generate similar results. Although \textit{Petalica Paint}, \cite{comicolorization}, and \cite{style2paintV2} are able to use reference images and color dots to colorize an image with plausible colors for each region, for comparison with our method, we did not use these hints.

\begin{figure*}[!ht]
\minipage[t]{0.47\textwidth}
  \includegraphics[width=\linewidth]{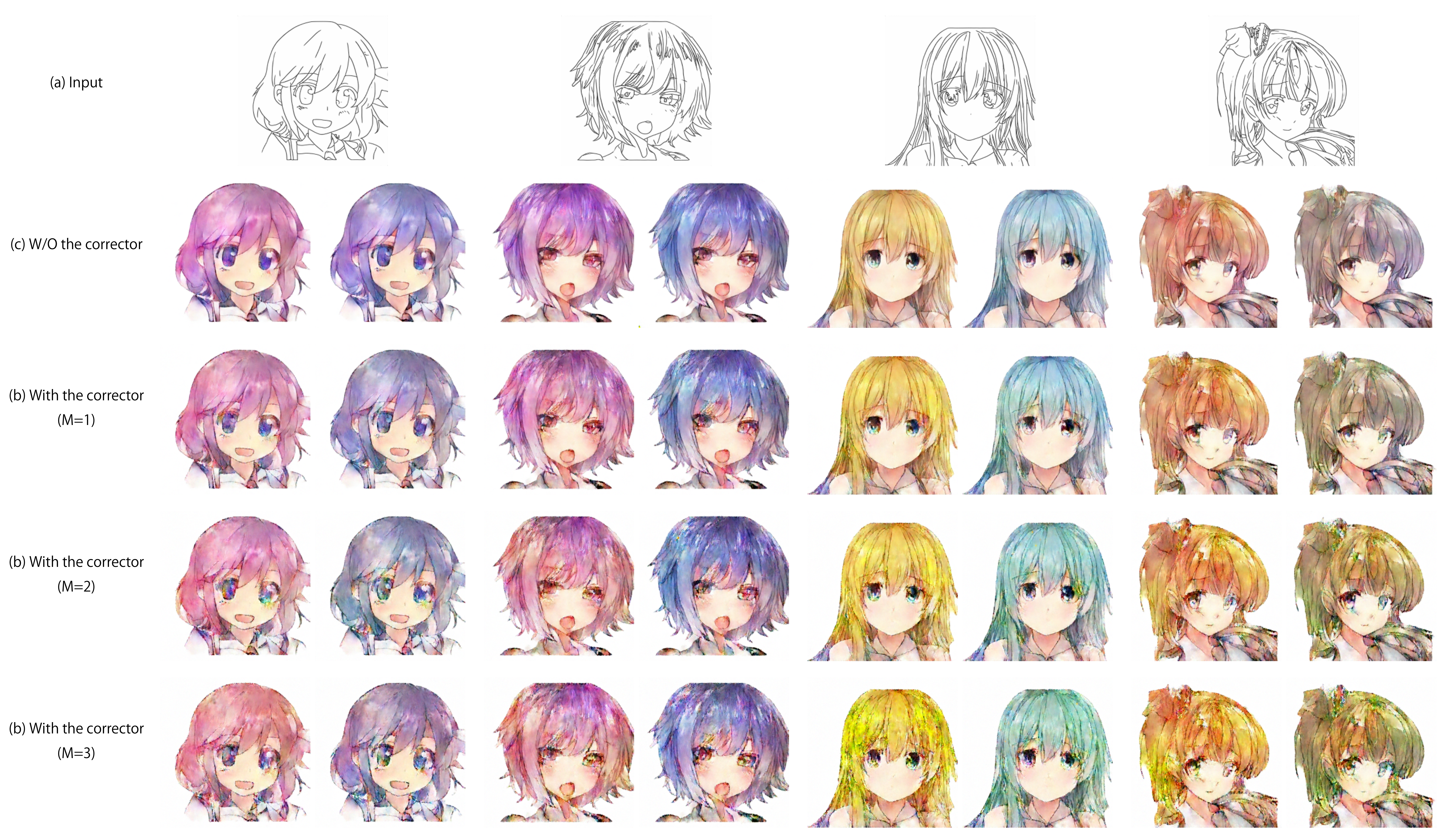}
  \caption{
  (a) Input images. 
  (b) Without the corrector. 
  (c) With the corrector ($M=1$). 
  (d) With the corrector ($M=2$). 
  (e) With the corrector ($M=3$). 
  It seems that larger values of M increase the noise in the generated image.
}
  \label{fig:corrector}
\endminipage\hfill
\minipage[t]{0.47\textwidth}
  \includegraphics[width=\linewidth]{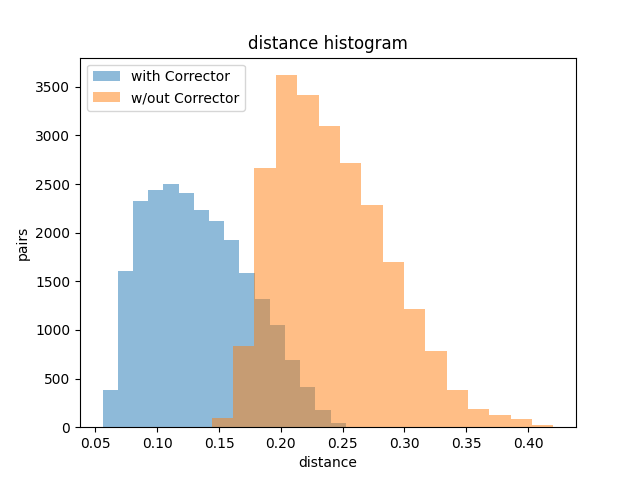}
  \caption{
  Similarity of generated images with and without the corrector. The similarity was measured by the distance as computed based on \cite{PerceptualSimilarity}. Images generated without the corrector had greater distances than images generated with it.
}
  \label{fig:corrector_dists}
  \endminipage\hfill
\end{figure*}

\Fref{fig:comparison}(i) and (j) show multiple candidate colorizations. However, these methods have some issues, as shown in detail on the right. The top row is by \cite{munit}, the middle row is by \cite{MSGAN}, and the bottom row is by our method. \cite{munit} and \cite{MSGAN} have some issues: they produce ghostly, washed-out images, or exhibit color bleeding outside the desired regions. 
These issues influence the quantitative evaluation in the next paragraph. In contrast, our results in the bottom row have no missing parts and no color bleeding.  \Fref{fig:comparison}(k) shows that our method  generated multiple and diverse colorization results, with 
\begin{wrapfigure}{ro}{0.6\linewidth}
\centering
  \includegraphics[width=\linewidth]{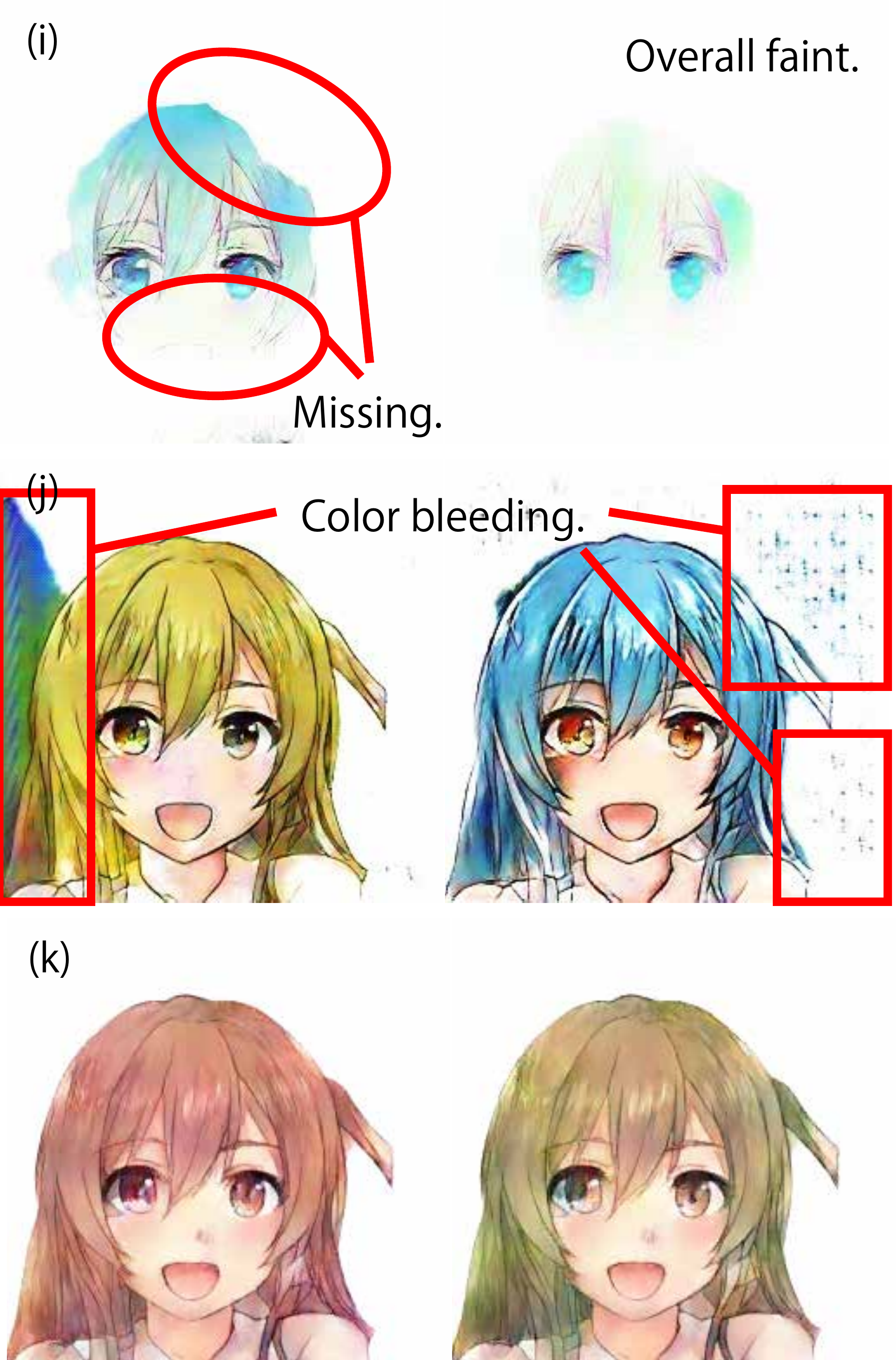}
  \captionsetup[figure]{labelformat=empty}
  \label{fig:comparison_closeup}
\end{wrapfigure}
pink, cyan, red, green, yellow, and blue color schemes. Our results are stable, the network is clear, and our network simply uses L1 loss as the loss. We attempted to perform the experiment using our dataset with a method inspired by \cite{SRflow, Glow}, however the loss ended up diverging. In addition, the size of the model trained using this method was very large. 

We next quantitatively evaluated our method using image similarity, focusing on the resulting distribution of generated images. We measured the diversity of the generated colorized images using a perceptual evaluation metric \cite{PerceptualSimilarity}. Their method computes the distance between two images as the output of a neural network:

\begin{equation}
d\left(x, x_{0}\right)=\sum_{l} \frac{1}{H_{l} W_{l}} \sum_{h, w}\left\|w_{l} \odot\left(\hat{y}_{h w}^{l}-\hat{y}_{0 h w}^{l}\right)\right\|_{2}^{2}.
\end{equation}
where $l$ is the layer index, and $\hat{y}^{l}, \hat{y}_{0}^{l} \in \mathbb{R}^{H_{l} \times W_{l} \times C_{l}}$ are features output by the neural network at each layer $l$. Their network scales the activations channelwise by a vector $ w^{l} \in \mathbb{R}^{C_{l}}$, and computes the L2 norms. We posit that if a method produces images that have greater average distances between them, as well as a wider distribution of distances, then that indicates that the method is able to generate more diverse colorized images.
We generated 216 colorized images using both \cite{pix2pix} and our method, and then computed the distance between every possible pair using \cite{PerceptualSimilarity}. 

In \Fref{fig:comparison_dists}, the blue bins are the number of pairs of each distance obtained by \cite{pix2pix}, the orange bins are those obtained by \cite{munit}, the green bins are those obtained by \cite{MSGAN}, and the orange bins are those obtained by our method. The distribution of the blue bins, corresponding to \cite{pix2pix}, is narrow and near the origin. The distribution of the orange bins, corresponding to \cite{munit}, looks wide and far from the origin at a glance, but the greater distances beyond 0.3 in the histogram are due to one of the images in the pair being poorly generated, as described in the previous paragraph. These poor results caused the greater distances between the images in the pairs. The green bins corresponding to \cite{MSGAN} suffer from similar problems. On the other hand, our method produces reasonably varied images without sacrificing image quality and stability.

\subsection{Influence of the Corrector}
We considered the influence of the corrector on the output images. Because the computation was split between the predictor and corrector, we were able to use the corrector repeatedly during generation.
As may be observed qualitatively in the output results, there was significant noise in the results generated with the corrector. In \Fref{fig:corrector}, (a) are the input images, (b) are the images generated without the corrector, and (c)–(e) are the images generated with the corrector, where $M$ is the number of corrector iterations. It is evident that larger values of $M$ increased the noise in the output image. For example, (e) seems to have more noise than (a) or (b). 

Additionally, we analyzed the effect of the corrector on the similarity of generated images. \Fref{fig:corrector_dists} shows the similarity of images generated using the corrector and those generated without it. The similarity was measured by the distance as calculated based on \cite{PerceptualSimilarity}. \Fref{fig:corrector_dists} shows that images generated without the corrector had greater distances than images generated with it. Therefore, we concluded that the corrector was not appropriate for colorizing images, and in the end we omitted the corrector for every result.


\subsection{Combining models}
\label{sec:channels}

\begin{figure*}[!ht]
\minipage[t]{0.47\textwidth}
  \includegraphics[width=\linewidth]{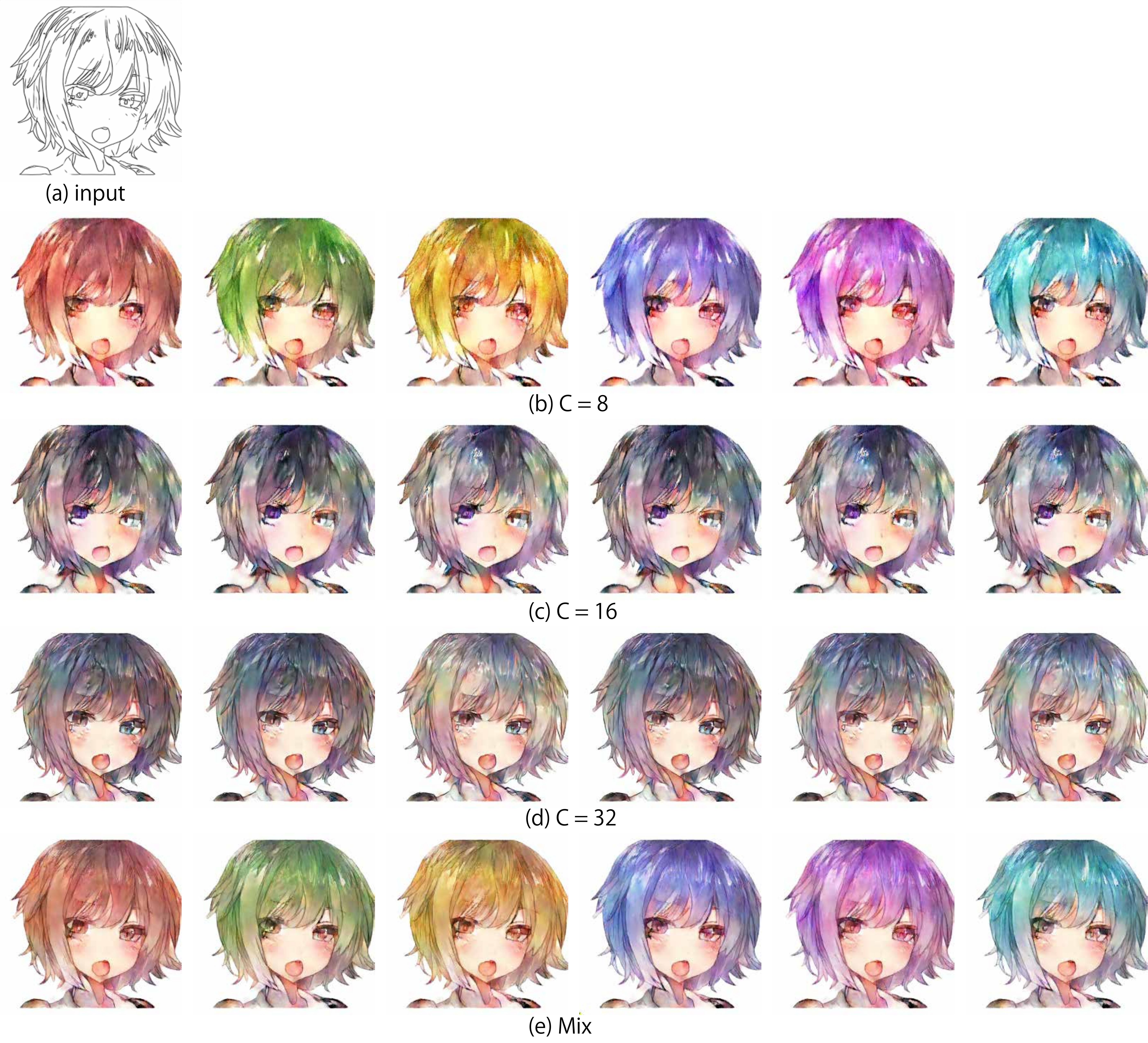}
  \caption{
  Differences in the results based on the number of channels in the middle layer. (a) Input. (b) Output generated by the C8 model. (c) Output generated by the C16 model. (d) Output generated by the C32 model. (e) Output generated by combining the C8 model and the C32 model. The results generated by the model with fewer channels were less sharp, but more vivid and diverse. To generate sharp, vivid and diverse results, we combined the C8 model and the C32 model as in (e).
}
  \label{fig:channels}
\endminipage\hfill
\minipage[t]{0.47\textwidth}
  \includegraphics[width=\linewidth]{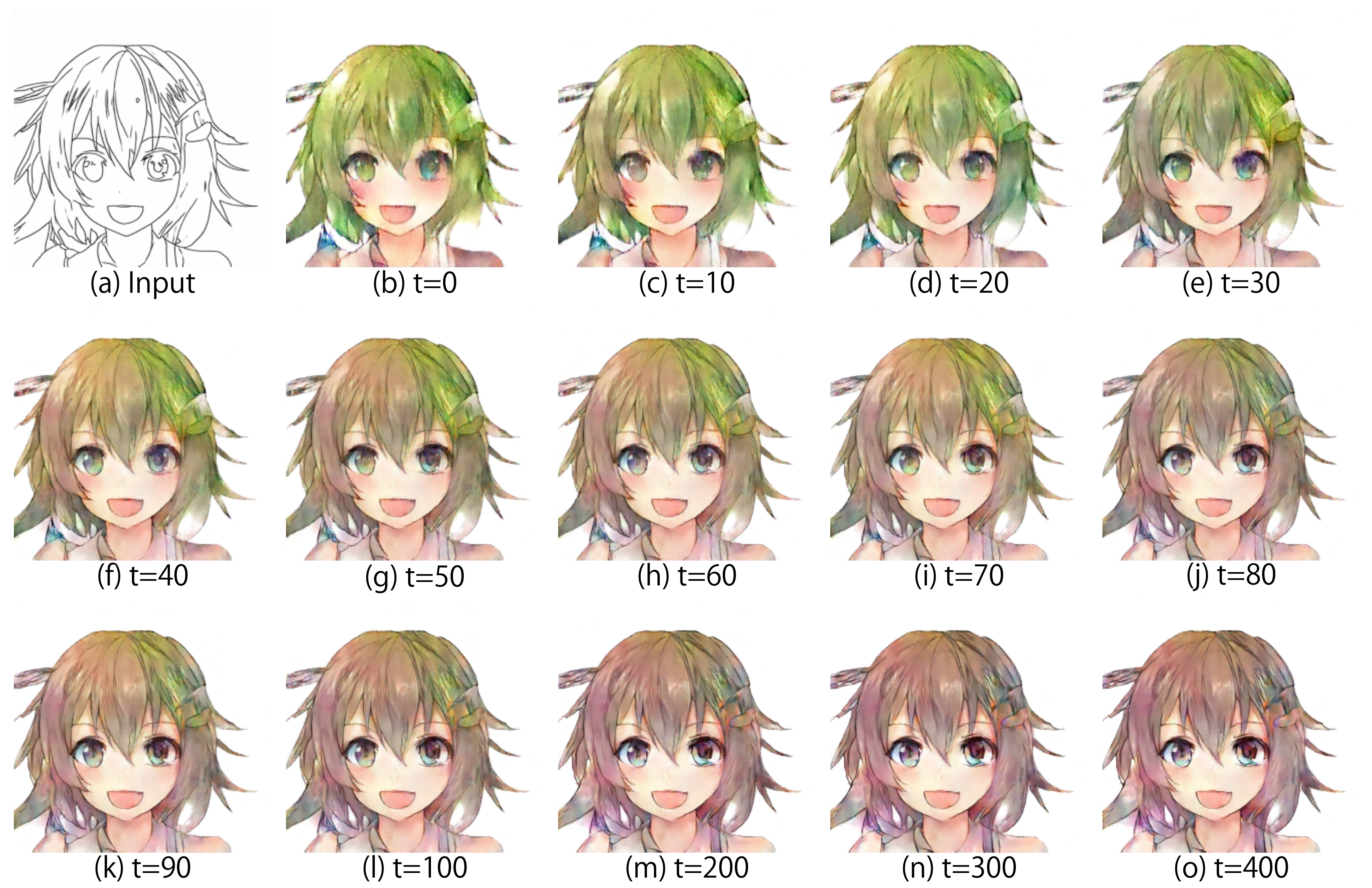}
  \caption{
  The effect of switching models on the results. (a) Input. (b)-(o) Outputs generated by the model when switched at the given iteration $t$. The total number of iterations T was 1000. As the timing of switching models approached the end of the iterations, the result became gradually more vivid, but also less sharp. Therefore, we concluded that switching at $t=30-50$ was ideal.
}
  \label{fig:switch_model}
  \endminipage\hfill
\end{figure*}

We experimentally found that the number of channels in the middle layer influenced the variation in colors, where the number of channels in the convolutional layers is determined by the parameter $C$ in \Fref{fig:network_main}. \Fref{fig:channels} shows the results of a comparative experiment using models with different numbers of channels. \Fref{fig:channels}(a) shows the input image, (b) was generated by a model with $ C=8 $, (c) shows the results generated by a model with $ C=16 $, and (d) shows the results generated by a model with $ C=32 $, which we will call the “C8 model”, "C16 model", and “C32 model” respectively for brevity.
While the images in \Fref{fig:channels}(b) are more vivid than (d),  and they use clearly different color compositions, the images in \Fref{fig:channels}(d) are sharper than (b). \Fref{fig:channels}(e) was generated by combining a C8 model and a C32 model, which we will call the “Mix model”.

\newpage
\begin{figure}[!ht]
\centering
  \includegraphics[width=\linewidth]{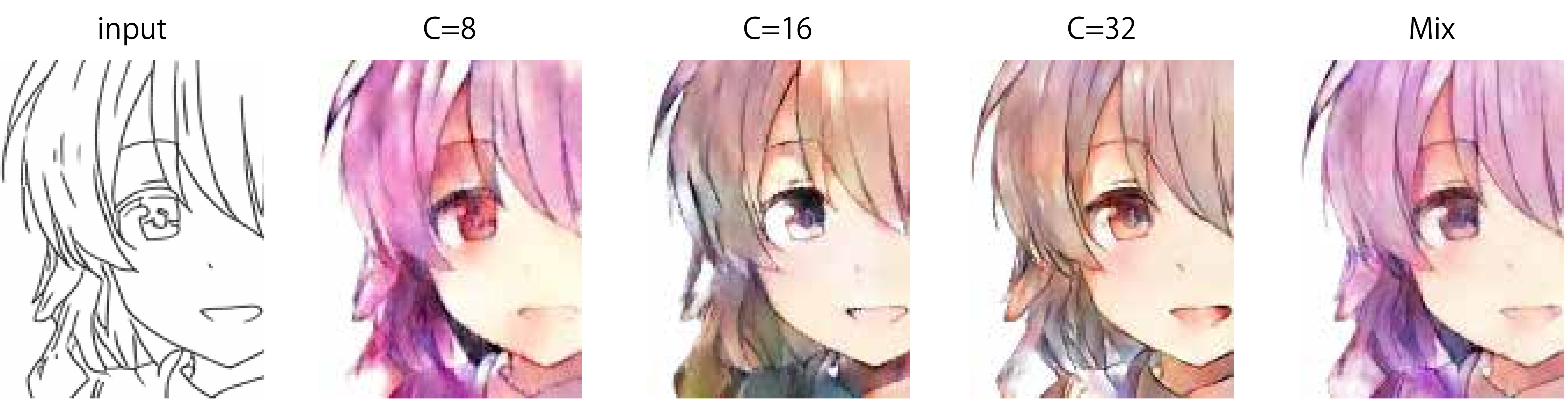}
  \captionsetup[figure]{labelformat=empty}
  \label{fig:ch_closeup}
\end{figure}

The close-up at the upper figure more clearly shows that the C32 model produces sharper output with less color bleeding than the C8 model. C16 model falls roughly between C8 model and C32 model in terms of vividness and sharpness. To generate both vivid and sharp results, we combined the models. By Mix model, we are able to obtain results that are not only vivid and diverse, but also sharp, as shown at the close-up figure. We used a C8 model for most of the generation process, switching to a C32 model near the end. We switched the model at $t=40$, or the 960th iteration out of 1000 of the reverse process.

\Fref{fig:switch_model} shows a comparison of the results of switching models at different points during generation. \Fref{fig:switch_model}(a) is the input image, while (b)-(o) were generated by a model that was switched at the given iteration $t$. As the timing of switching models approached the end of the iterations, the result became gradually more vivid, but less sharp. Therefore, we finally concluded that switching at $t=30-50$ was ideal.

\subsection{Other Results}
\begin{figure*}[t]
  \centering
  \includegraphics[width=\linewidth]{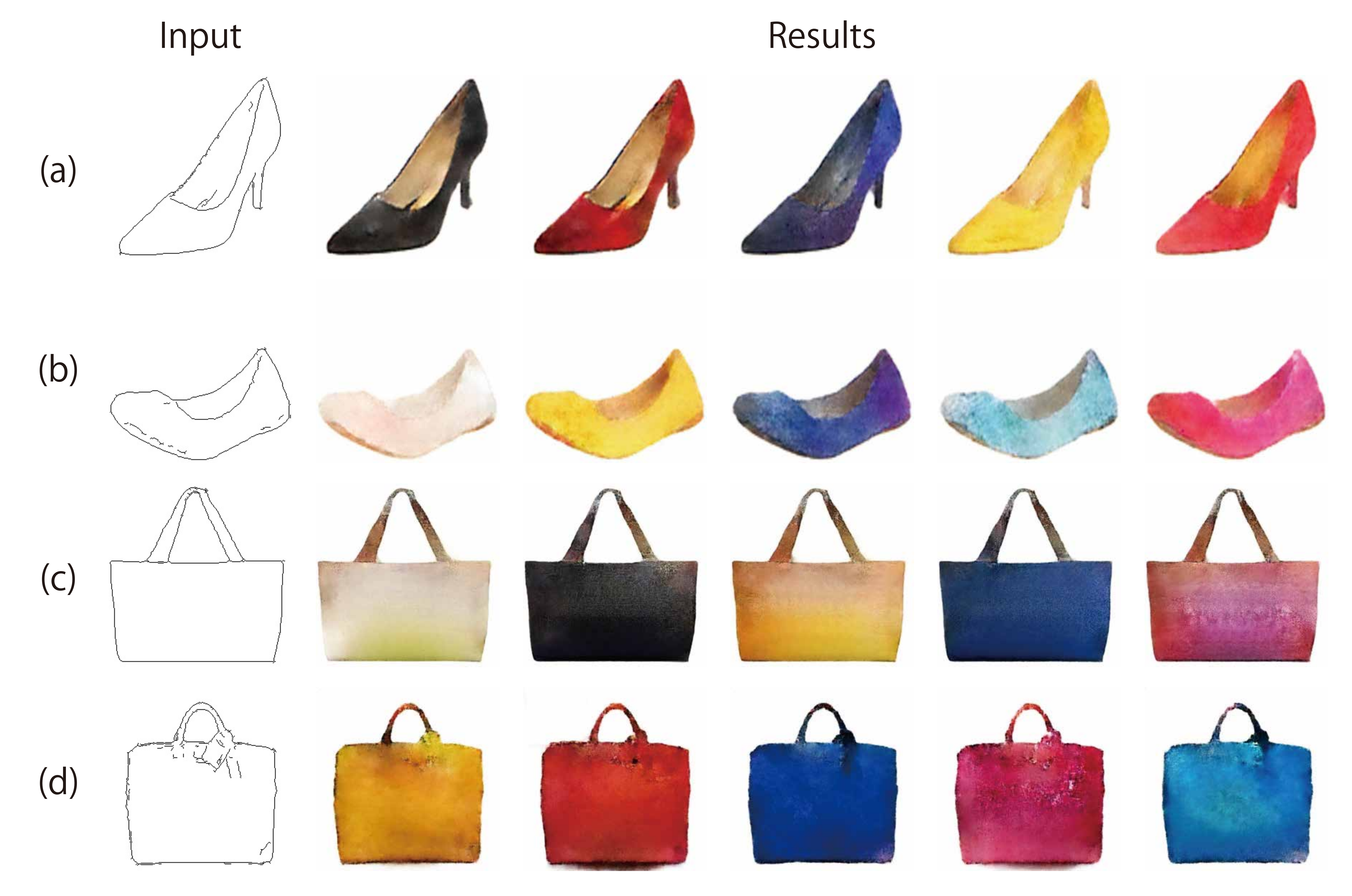}
  \caption{
  Results using the pix2pix datasets, with the input image on the left.
(a) and (b) are results using the edges2shoes dataset, and
(c) and (d) are results using the edges2handbags dataset.
Each row shows multiple and diverse colorizations.
These results show that our method can generate multiple colorizations, even for other datasets. Moreover, we conclude that our method can generate multiple and diverse results stably using a relatively small training dataset.
}
  \label{fig:sm_result1}
\end{figure*}

We experimented using other datasets to confirm whether our method is effective in other domains.
We selected 450 images from the edges2handbag and edges2shoes datasets in \cite{pix2pix}.
We also use both the C8 model and the C32 model, and we switched the model at $t=40$, or the 960th iteration out of 1000 of the reverse process.
\Fref{fig:sm_result1} shows the results using the pix2pix datasets, with the input image on the left.
(a) and (b) are results using the edges2shoes dataset, and
(c) and (d) are results using the edges2handbags dataset.
Each row shows multiple and diverse colorizations.
These results show that our method can generate multiple colorizations, even for other datasets. Moreover, we conclude that our method can generate multiple and diverse results stably using a relatively small training dataset.

\subsection{Practical Applications}
\label{sec:application}

\begin{figure*}[!ht]
\minipage[t]{0.47\textwidth}
  \includegraphics[width=\linewidth]{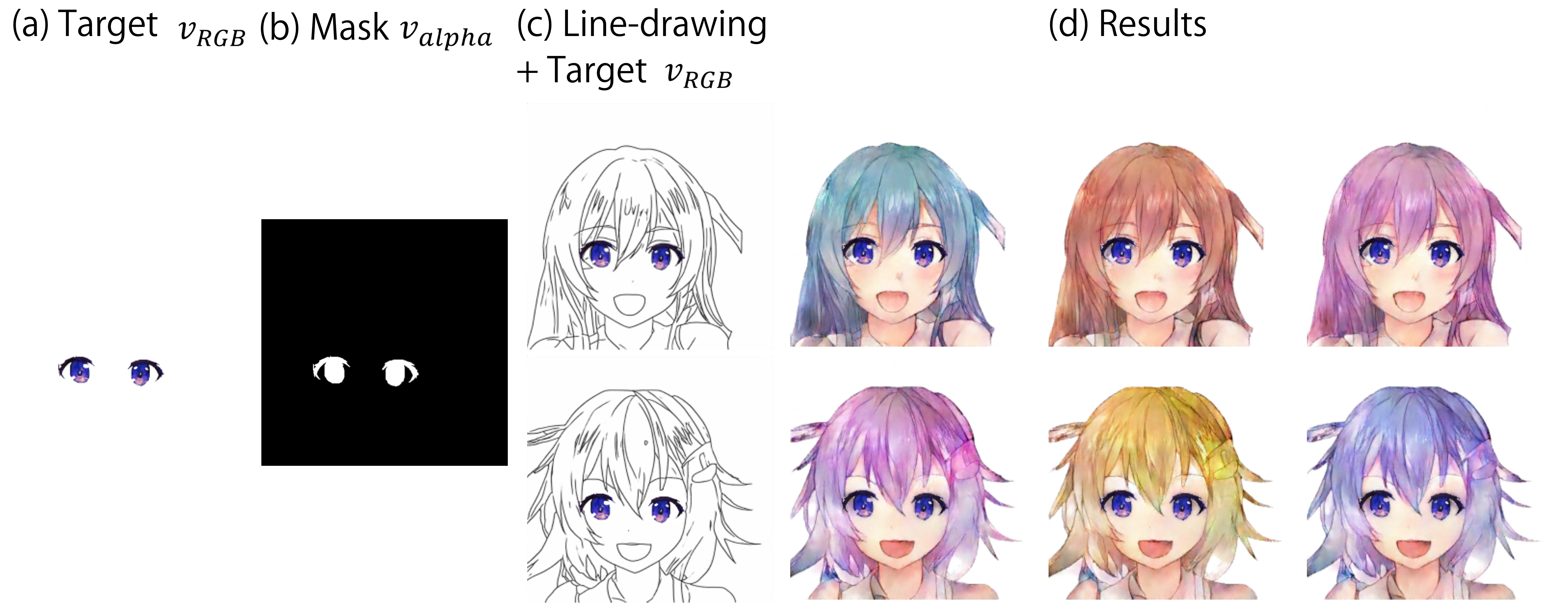}
  \caption{
(a) Target colored region $ v_{\text {RGB}} $ (b) The mask $ v_{\text {alpha}}$. (c) Input line-drawing image and the target $ v_{\text {RGB}}$. (d) The results. Our method demonstrated the capacity to combine a target region with an input line-drawing by the adding the step given in \Eref{eq:image_synthesis}, allowing our method to simultaneously perform image completion and colorization.
}
  \label{fig:matting}
\endminipage\hfill
\minipage[t]{0.47\textwidth}
  \includegraphics[width=\linewidth]{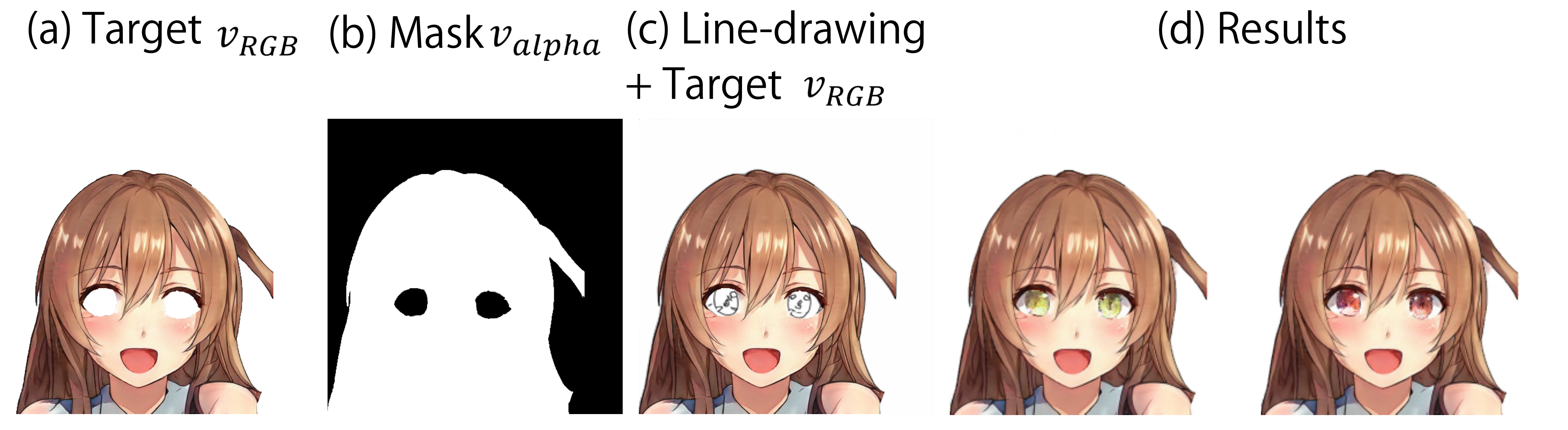}
  \caption{
(a) Target colored region $ v_{\text {RGB}} $ (b) The mask $ v_{\text {alpha}}$. (c) Input line-drawing image and the target $ v_{\text {RGB}}$. (d) The results. Our method demonstrated the ability to fill the blank region with the addition of the step in \Eref{eq:image_synthesis}. In other words, our method can also be used as an image synthesis technique, such as for image inpainting.
}
  \label{fig:inpainting}
  \endminipage\hfill
\end{figure*}

Given a partial colored image and a line-drawing around it, our method is able to colorize the rest of the image. To do so, we add a step that modifies the value of ${x}_{t}$ at the start of the outermost for loop (i.e. before the step on line 4) in Algorithm 2 as follows.
\begin{equation}
{x}_{t} \leftarrow {x}_{t} \odot (1 - v_{\text {alpha}}) + v_{\text { RGB }} \odot v_{\text {alpha}}, 
\label{eq:image_synthesis}
\end{equation}
where $v = (v_{\text { RGB }}, v_{\text {alpha}})$ represents the region of the partial colored image, with $v_{\text {RGB }}$ being the RGB channels of $v$, and $v_{\text {alpha }}$ being the alpha channel of $v$.

Our method is able to complete and colorize the image if the user inputs a line-drawing around the partial colored image. \Fref{fig:matting} shows examples of image synthesis using this extra step. \Fref{fig:matting}(a) is the target partial colored image. To clarify the purpose of the task, \Fref{fig:matting}(c) shows the input line-drawings combined with the target image $ v_{\text {RGB}}$. \Fref{fig:matting}(b) is the mask $ v_{\text {alpha}}$, and \Fref{fig:matting}(d) are the results of synthesis. These colorization results show that the region around the eyes was plausibly synthesized. Our method can transfer a region of the target image to the constructed image by adding the step in \Eref{eq:image_synthesis}, and can therefore be used for simultaneous image synthesis and colorization, such as in the task of image completion with structural annotations.

In the same way, our method is able to fill a blank region and colorize the image if the user inputs a partial line-drawing.
\Fref{fig:inpainting}(a) is the target colored image which has a blank which is the region inside the character’s eyes. 
\Fref{fig:inpainting}(c) shows the target image $ v_{\text {RGB}}$ combined with the input line-drawing. \Fref{fig:inpainting}(b) is the mask $ v_{\text {alpha}}$, and \Fref{fig:inpainting}(d) are the results of synthesis. \Fref{fig:inpainting}(d) shows that the blank region inside the character’s eyes was inpainted. Moreover, the eye regions were painted differently for each candidate. Our method demonstrated the ability to inpaint a blank region by adding the step in \Eref{eq:image_synthesis}, making it suitable for the inpainting task.

\section{Conclusion}
In this study, we have proposed an image colorization method based on diffusion probabilistic models, which are a class of latent variable models inspired by models used in nonequilibrium thermodynamics. We experimented with diffusion models and compared our results to GAN-based and flow-based colorization methods in terms of their ability to explicitly account for the ill-posed nature of the problem of image colorization and generate multiple and diverse candidate colorized images. We experimentally found that our method can generate multiple and diverse colorization candidates which avoid mode collapse by combining a C8 model and a C32 model. Additionally, we examined the characteristics of GAN-based methods and flow-based methods on the same task in comparison to our method. GAN-based methods can generate multiple colorization candidates, but sometimes generate results with missing details, and also require complex tuning of the network, loss, and hyper-parameters. Training of flow-based methods is often unstable, and in our experiments, the loss of this method ended up diverging. We perform experiments to test the effect of the corrector, and we confirm that it could be applied to our algorithm, however, we concluded that the corrector was not appropriate for the colorization task. Moreover, we developed techniques for conditional image generation that do not require retraining, and also demonstrated some practical applications for image completion and inpainting tasks.

\newpage

\clearpage
\bibliographystyle{ACM-Reference-Format}
\bibliography{ACM-Reference-Format}

\end{document}